# Reducing Formal Context Extraction: A Newly Proposed Framework from Big Corpora

Bryar A. Hassan[1,2], Shko M. Qader[3], Alla A. Hassan[4], Joan Lu[5], Aram M. Ahmed[1], Jafar Majidpour[6], Tarik A. Rashid[1]

[1]Computer Science and Engineering Department, School of Science and Engineering, University of Kurdistan Hewler, Erbil, Iraq

[2]Department of Computer Science, College of Science, Charmo University, Chamchamal 46023, Sulaimani, KRI, Iraq.

[3]Information Technology Department, Kalar Technical Institute, Garmian Polytechnic University, Sulaimani, Iraq.

[4]Department of Database Technology, Computer Science Institute, Sulaimani Polytechnic University, Sulaimani, Iraq.

[5]Department of Computer Science, School of Computing and Engineering, University of Huddersfield, Queensgate, Huddersfield HD1 3DH, UK

[6]Department of Computer Science, University of Raparin, Rania, Iraq

Corresponding email: bryar.ahmad@ukh.edu.krd

**Abstract**

Automating the extraction of concept hierarchies from free text is advantageous because manual generation is frequently labor- and resource-intensive. Free result, the whole procedure for concept hierarchy learning from free text entails several phases, including sentence-level text processing, sentence splitting, and tokenization. Lemmatization is after formal context analysis (FCA) to derive the pairings. Nevertheless, there could be a few uninteresting and incorrect pairings in the formal context. It may take a while to generate formal context; thus, size reduction formal context is necessary to weed out irrelevant and incorrect pairings to extract the concept lattice and hierarchies more quickly. This study aims to propose a framework for reducing formal context in extracting concept hierarchies from free text to reduce the ambiguity of the formal context. We achieve this by reducing the size of the formal context using a hybrid of a WordNet-based method and a frequency-based technique. Using 385 samples from the Wikipedia corpus and the suggested framework, tests are carried out to examine the reduced size of formal context, leading to concept lattice and concept hierarchy. With the help of concept lattice-invariants, the generated formal context lattice is compared to the normal one. In contrast to basic ones, the homomorphic between the resultant lattices retains up to 98% of the quality of the generating concept hierarchies, and the reduced concept lattice receives the structural connection of the standard one. Additionally, the new framework is compared to five baseline techniques to calculate the running time on random datasets with various densities. The findings demonstrate that, in various fill ratios, hybrid approaches of the proposed method outperform other indicated competing strategies in concept lattice performance.

# 1. Introduction

In recent years, the need to organize and extract meaningful information from large and unstructured data sources has become increasingly crucial across various fields, such as knowledge management, e-government, and healthcare. Concept hierarchy extraction, a technique that structures data into hierarchical relationships, plays a key role in this process. Formal Concept Analysis (FCA) has emerged as one of the most effective methods for deriving concept lattices from data [1]. It offers a robust mathematical framework for structuring relationships between objects and their attributes. However, existing FCA methods face challenges when applied to large datasets, particularly in balancing concept lattice reduction with the preservation of important hierarchical structures [2].

Moreover, the Semantic Web, an enlarged web of machine-readable data, offers a program to process data by machine directly or indirectly [3], [4]. The Semantic Web, an extension of the most current web, can help automate services based on semantic representations and provide meaning to content on the World Wide Web. The Semantic Web uses structured ontologies to arrange the underlying data and provide a thorough and portable interpretation of computer systems [5]. Information Systems frequently make use of ontologies, which are a crucial component of the Semantic Web.

Additionally, the growth of ontologies necessitates speedy and effective ontology creation to ensure the success of the Semantic Web [6]. Nevertheless, manually creating ontologies is a laborious procedure, with the major problems being gathering information, slow growth, maintenance challenges, and integrating many ontologies for various practical applications and fields. A way to get around the difficulty of quickly learning new objects and developing ontologies is through ontology learning. In the knowledge extraction subtask of ontology learning, ontologies are created by automatically or semi-automatically extracting pertinent ideas and connections from a corpus or other data sets [7] that describes the current methodology for automatically extracting concepts and concept hierarchies from corpora. Based on FCA, this method creates a concept lattice that can be transformed into a partial order that makes up a hierarchy of concepts. Conversely, the formal context may contain some incorrect and uninteresting pairings because not all the derived pairs are accurate [8]. The formal context was built using a sizable free text corpus, which can provide inaccurate output from the parser. Nevertheless, extracting the concept lattice from it may be time-consuming, depending on the volume and complexity of the formal context data [9]. By eliminating the unnecessary and inaccurate pairings from the formal context, it may be possible to derive the concept lattice more quickly. Despite this, there is a lack of efficient methods to automatically refine and filter formal contexts to address the inaccuracies and irrelevant pairings, which hinders the scalability and reliability of concept lattice generation in ontology learning.

Hence, the primary contributions of this research include a new and original framework for automatically extracting concept hierarchies from a free text that departs from the existing framework. Otherwise, the current system includes many steps for extracting idea hierarchies from the text. One of the major processes in this framework is the extraction of word pairs or phrase dependencies, particularly the vital and interesting pairings. This is because not all couplings are correct or fascinating. The word combinations are then weighted using a statistical measure, and those more significant than a certain threshold is converted into formal contexts to which FCA is applied. The idea hierarchies are then produced using the concept of the lattice. On the other hand, the suggested framework uses the same techniques as the earlier one. Still, it omits unnecessary and inaccurate pairings from the formal context, shrinking the formal context and speeding up deriving the concept lattice [10]. To do this, two different strategies are used in the formal environment, resulting in a reduced concept lattice and hierarchy. They are evaluated to see if the concept lattices produced by the frameworks are isomorphic to one another. In addition, this study proposes a hybrid approach that addresses these limitations by combining WordNet-based semantic analysis and frequency-based statistical reduction [11], [12]. The WordNet-based method enhances semantic accuracy by grouping related concepts, while the frequency-based approach improves computational efficiency by filtering out infrequent and irrelevant concepts. Together, these methods allow for more effective formal context reduction, leading to faster and more accurate concept lattice generation. Specifically, the two main methods for minimizing formal contexts are the WordNet-based and frequency-based techniques. Only the pairings that score higher than a given threshold are translated into a formal context where FCA is applied. The former is used to weigh the pairings based on some statistical metric. With WordNet serving as a lexical database for the English language, the latter is utilized to weigh the pairwise comparisons. Determining the formal context size to extract a distinct and meaningful concept hierarchy from text corpora is the primary motivation for conducting this project. Significantly, when the size of the formal context is lowered, the complexity of the concept lattice and concept hierarchy is also decreased. This approach has three advantages: it could take less time to create idea hierarchies; producing meaningful concept hierarchies might be aided by utilizing two separate approaches; and shrinking the size of concept hierarchies gets rid of useless and false information.

Despite the advancements in FCA and concept hierarchy extraction, several research gaps remain:

- Scalability: Traditional FCA-based methods, such as AddIntent and NextClosure, struggle with scalability when handling large and complex datasets [13].
- Semantic Limitations: Many existing techniques lack mechanisms for integrating semantic relationships between concepts, which can result in poor concept grouping and inaccurate hierarchies [14].
- Efficiency vs. Accuracy Trade-Off: Many methods focus either on lattice reduction or accuracy but fail to balance both effectively [15].

It needs to be mentioned that this study makes the following contributions:

- We introduce a novel hybrid framework that integrates WordNet-based semantic analysis with frequency-based filtering to improve formal context reduction.
- The proposed method demonstrates significant improvements in computational efficiency by reducing the size of the concept lattice while maintaining high structural accuracy.
- We validate the generalizability of the framework across multiple domains, including healthcare, e-government, and educational technology, by applying it to diverse datasets.
- We present a detailed comparison of our method with existing techniques, demonstrating its superior performance in both semantic coherence and computational efficiency.

The following sections of this study are organized: The earlier studies on getting concept hierarchies from corpora and decreasing concept lattices in the literature are summarized in Section 2. The formal context size that might arise in concept lattice when concept hierarchies are formed from free text using two different ways is reduced utilizing a framework proposed in Section 3. On 385 datasets, an experiment is run in Section 4 to compare the reduced concept lattices to the original ones. Section 5 is experimentally compared to its comparable algorithms and the recommended approach, including result analysis, performance evaluation, and some limitations of the proposed framework. The last remarks are then discussed, commenting on the flaws of this work to demonstrate how this study will help scholars and offering suggestions for future research.

## 2. Related Work

The development of automated ontologies is one of Semantic Web's current research projects. Since it takes topic specialists time and resources to manually develop the ontology, ontology learning is seen as a time- and resource-intensive process [16]. Therefore, it would be advantageous to partially or entirely support this strategy for constructing ontologies gradually or semi-automatically. Ontology learning is the semi-automatically extracting relationships and important concepts from text corpora or other data sources.

### 2.1. Formal Concept Analysis and Lattice Reduction

The development of ontologies over the past ten years has been supported by several techniques, from machine learning, natural language processing, knowledge representation, information retrieval, and data mining [16]. Computing techniques like data mining, machine learning, and knowledge retrieval may be employed to identify domain names, meanings, and connections. On the other hand, natural language processing makes a major contribution to almost every level of ontology learning layer cake by offering linguistic tools. FCA is a fascinating statistical technique to construct concept hierarchies. This approach is based on the notion that characteristics are linked to the qualities of the objects they are associated with. It uses the matrix property of an item as input to locate all the natural attributes and object groupings collectively. It creates a lattice with definitions and traits shaped like a hierarchy. The widespread use of FCA is undoubtedly not a fresh concept. Lexical tuning, lexical-semantic analysis, and language structure analysis are just a few of the FCA's potential applications [17]. A novel technique for autonomously deriving idea hierarchies from domain-specific literature was proposed in a recent work [18], with a focus on FCA. On two datasets, the method outperformed a hierarchical agglomerative clustering algorithm and Bi-Section-K-Means.

Several information metrics were also examined to establish the value of an attribute/object pair. The conditional probability was found to perform well, in contrast to other more sophisticated information metrics. The generic method for automatically creating definition hierarchies from text corpora was provided in the same study. There are several stages to automating idea hierarchies with Systematic idea Analysis. Before making a parse tree, the text is first recognized as part of speech (POS) and parsed for each phrase. The parse trees are then used to extract the links between the verb and the prepositional phrase, verb and object, and verb and subject. Particularly, pairings of the verb and the head of the object, topic, or prepositional phrase are retrieved. After being lemmatized, the verb and its head are converted to their basic form. Figure (1) shows that word pairs may be eliminated from text corpora using NLP components.

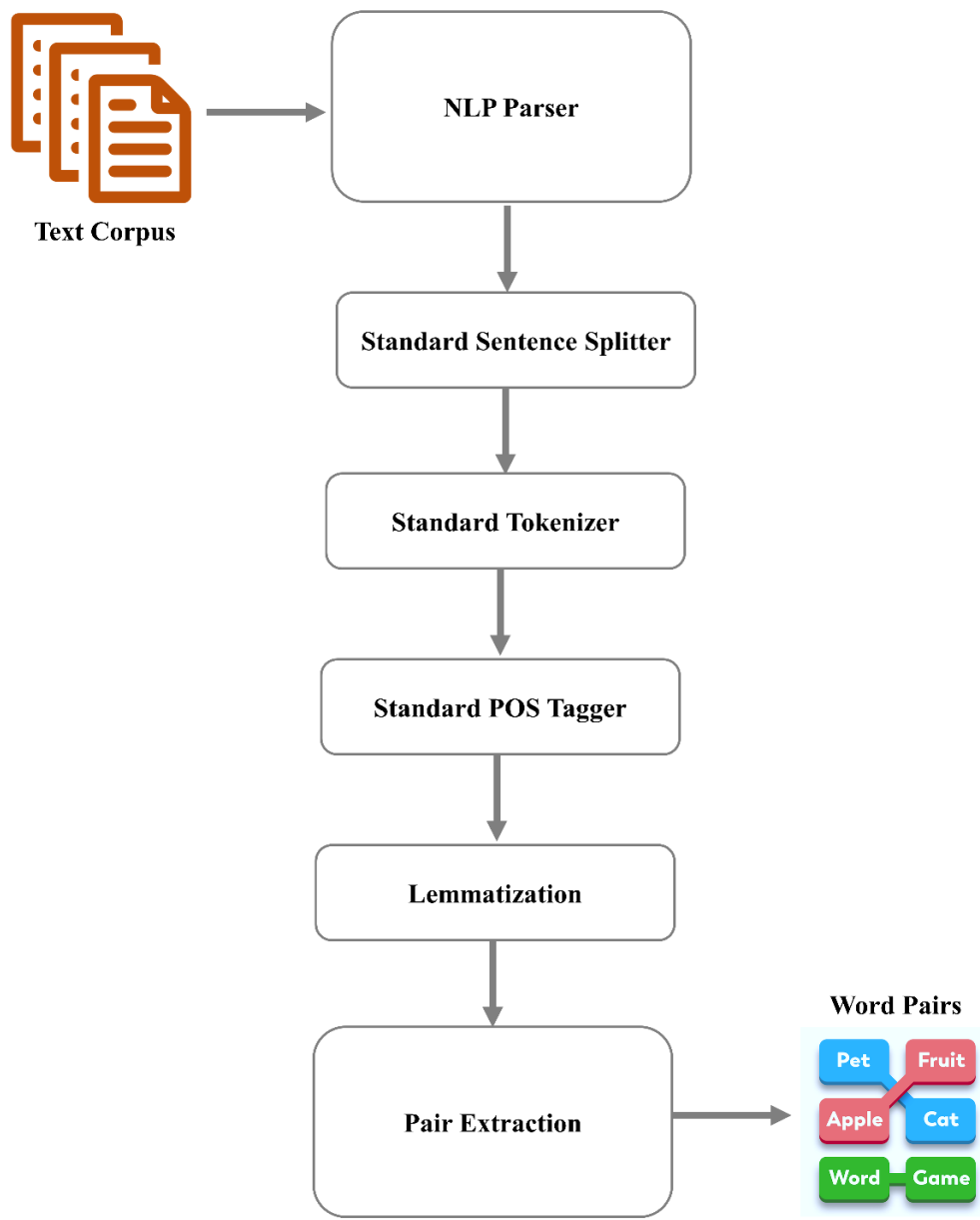

Figure (1): The main components of NLP

Pair collection is also smoothed to address data sparsity, which suggests that pairs that disappear from the corpus are there based on the frequency of other pairs. The pairs are then legally subjected to the FCA framework. Figure (2) depicts the procedure for autonomously constructing concept hierarchies from text corpora.

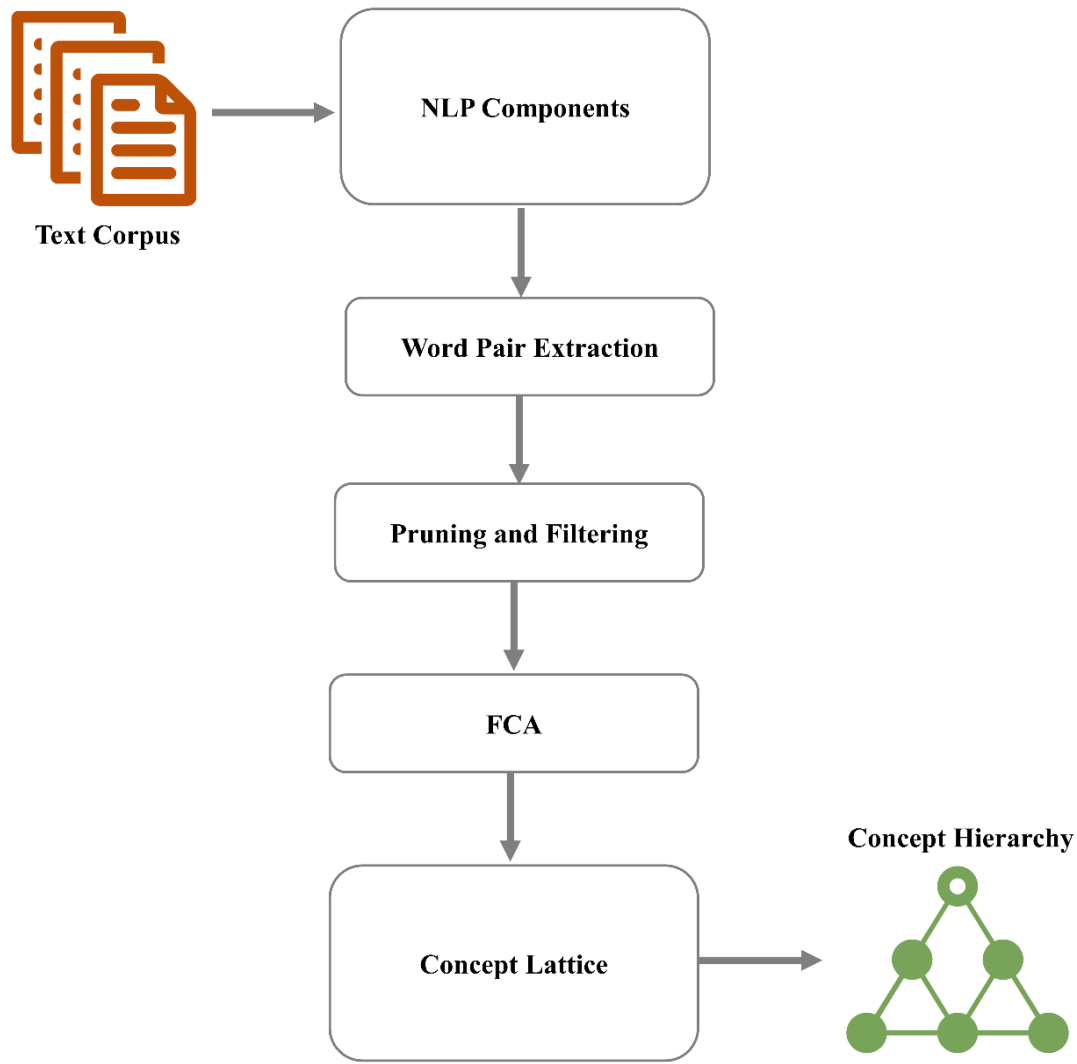

Figure (2): The existing framework for automatically building concept hierarchies (adapted from [18])

The recovered pairs might not always be unique, resulting in inaccurate concept hierarchies that generate appropriate and fascinating pairings and lower the formal context size. A reduced formal context could make it easier to extract the concept lattice by removing unimportant and inaccurate pairings. Even though Dias and Vieira [19] assess various ways for concept lattice reduction, numerous methodologies have been established for the reduced structures of concept lattice. The authors of the same review classified the three concepts of lattice reduction techniques utilized in the literature. The first group focuses on duplicate data removal. The more straightforward defining lattice of duplicate information remover approach may be more suitable for applications needing direct user contact since it frequently has fewer downsides. This might happen by clearly understanding the lattice or using the implications of the current formal setup. This more practical strategy makes the definitions' area smaller than prior methods to employ the second group. Due to this trait, the consistency of the produced definition lattice differs from the conventional one. The approaches can be applied in a range of formal contexts since they are generally utilized in formal settings and have a level of sophistication. However, the final concept lattice could be poor quality and differ significantly from the original. The last group uses key factors, such as adding an objective function, to eliminate seemingly pointless routes on the concept lattice. This strategy prunes the space of concepts. As described by [20], constructing iceberg concept lattices is a well-known method in this area. The disadvantage of this strategy is that crucial formal structures are missed. In addition, other approaches create a list of all formal notions and choose the ones that apply using a set of parameters [21], [22], [23]. Any method based on this strategy may be costly since the search space is searched.

Seven criteria were selected from the most significant ones in the literature examining and classifying procedures in the same study [19]. FCA was used for the analysis, and a formal context was used to present the findings. All reduction methods have been shown to change the occurrence relationship and cause variable levels of descriptive loss. The first half of the techniques build on the collection of principles by focusing primarily on duplicate information removal approaches. However, the second half of the methods are expanding the FCA. They employ heuristic tactics one-fourth of the time and accurate algorithms three-quarters of the time. Although most simplification techniques start with the formal sense, generate a non-isomorphic non-subset of the original concept lattice, and use exact algorithms, some rely on historical data. Last, selection processes start with a formal framework, don't rely on historical data, and use precise algorithms. Simplifying and employing chosen ways, which might significantly minimize the area served, would surely result in the most significant savings. Simplifier techniques, however, are inherently harmful since they can dramatically alter the collection of formal definitions. The process leading to such updates must consider the model lattice's essential components. Selector approaches are exciting because they condense the idea space, but moving through it carefully is crucial to finding the proper lattices.

Recently, [2] conducted research to provide two approaches to conceptualize the concept of derivation. Firstly, a novel framework was presented employing the adaptive evolutionary clustering algorithm star (ECA*) to exclude the defective and uninteresting pairings from the formal context and to reduce the size of the formal context, which would result in a less time-consuming concept lattice. Secondly, the most recent approach for creating idea hierarchies from free text using FCA was examined. In the same study, an experiment was conducted to contrast the results of the reduced concept lattice with those of the original lattice. When seen via the lens of the experiment and result analysis, the end-result lattice is a homeomorphism to the conventional one while keeping the structural relationship between the two concept lattices. The two lattices' resemblance preserves the quality of the thought hierarchies that 89% of them create, in contrast to the basic one. The quality of the created concept hierarchies is promising, with an 11% information loss across the two idea hierarchies. The time it took for adaptive ECA* and its competing approaches to run on random datasets with different fill ratios were compared experimentally in the final step. The results demonstrate that adaptive ECA* outperforms other approaches (low, medium, and high) for the random dataset with three densities.

Most recently, Marchetti et al. (2023) propose a novel approach that focuses on enhancing lattice reduction techniques through deep learning-based geometric models [24]. The method integrates self-supervised learning mechanisms to achieve significant efficiency in reducing large-scale concept lattices while maintaining accuracy.

By leveraging neural networks, this approach demonstrates improved performance in handling semantic integration and concept hierarchies derived from large-scale linked data. This work sets the foundation for scalable and efficient concept lattice generation.

Similarly, Shao et al. (2024) extend the focus on scalability and real-time processing by introducing a methodology for constructing multi-granularity generalized one-sided concept lattices [25]. Their framework addresses the challenges of large datasets by using incremental lattice updates and hybrid FCA. Designed to process dynamic data such as IoT sensor inputs, this method emphasizes adaptability and practical applicability, ensuring continuous knowledge extraction and hierarchical organization.

Building on the need for efficient formal context reduction, Dudyrev and Ignatov (2023) present innovative techniques for optimizing formal context size using genetic algorithms and heuristic pruning [26]. By reducing computational overhead and preserving hierarchical relationships, the proposed approach demonstrates utility across open datasets, particularly in domains such as healthcare and finance.

In addition, Behrisch and Renkin (2023) contribute to the dynamic aspect of concept hierarchy extraction through their framework for computing witnesses for centralizing monoids on a three-element set [26]. Focusing on real-time data environments, their approach uses dynamic FCA and real-time data processing to enhance the accuracy and speed of lattice derivation. This solution is particularly suited for adaptive learning systems, such as those used in educational data streams, where continuous updates to concept hierarchies are required.

Together, these studies address critical challenges in FCA, including scalability, semantic integration, and computational efficiency. By leveraging techniques such as deep learning, incremental updates, genetic algorithms, and real-time processing, they advance the capabilities of concept lattice generation and formal context optimization for modern large-scale and dynamic datasets.

**2.2. Critical Analysis of Existing Work**

While many existing methods have demonstrated success in extracting and reducing concept hierarchies, they face several limitations:

- Semantic Understanding: Techniques that rely solely on semantic embeddings (e.g., Word2Vec, BERT) capture contextual relationships but often fail to deliver a structured output suitable for FCA, necessitating additional steps for ontology construction.
- Scalability: Traditional FCA-based methods, including AddIntent and NextClosure, are limited by their scalability, particularly when applied to large, complex datasets. These methods often generate excessive lattice complexity, requiring additional steps to prune and simplify the hierarchies.
- Efficiency: While methods like ECA* offer efficiency improvements, they still face challenges in balancing reducing lattice size and preserving critical conceptual relationships.

Our framework addresses these limitations by integrating semantic and statistical reduction methods. WordNet ensures that semantic relationships between concepts are captured and merged where appropriate, while the frequency-based method eliminates infrequent and less relevant concepts, improving both the efficiency and accuracy of the generated concept lattices. Additionally, our framework demonstrates superior performance in terms of computational efficiency, as evidenced by the results of our comparative experiments with existing techniques.

## 2.3. Summary of Related Work

Table (1) summarizes the key findings, techniques, and data sources used in recent studies on concept hierarchy extraction and FCA-based lattice reduction.

Table (1): Summary of related work

| Study | Key Findings | Techniques Used | Datasets |
|---|---|---|---|
| [27], 2004 | Incremental lattice construction, slow performance on large datasets | AddIntent, incremental construction | UCI Machine Learning Repository |
| [28], 2011 | Stability of formal concepts in large datasets | Stability analysis, lattice reduction | Formal context from various domains |
| [29], 2015 | Concept lattice reduction by removing redundant information | Duplicate data removal, concept pruning | Real-world datasets (various domains) |
| [30], 2019 | Semantic similarity for linked data using WordNet | WordNet-based similarity, linked data | Linked data, WordNet |
| [18], 2020 | Automated construction of concept lattices using FCA | FCA-based concept lattice construction | Text corpora (domain-specific) |
| [2], 2021 | Improved execution time through clustering in FCA reduction | Adaptive ECA*, clustering | Text corpora, Wikipedia corpus |
| [24], 2023 | Enhanced lattice reduction using deep learning for semantic integration | Deep learning, semantic analysis | Large-scale linked data |
| [25], 2023 | Efficient FCA-based ontology learning for real-time applications | Hybrid FCA, incremental lattice updates | IoT sensor data |
| [31], 2024 | Optimization of formal context using evolutionary algorithms | Genetic algorithms, heuristic pruning | Open datasets (healthcare, finance) |
| [26], 2024 | Dynamic concept hierarchy extraction for adaptive learning systems | Dynamic FCA, real-time data processing | Educational data streams |

## 3. The Proposed Framework

Two fundamental methods can be utilized formally and in the experiment. The first method is a WordNet-based method that is linguistic. The second method is a statistical method known as a frequency-based method. Only the pairings that score higher than a given threshold are translated into a formal context where FCA is applied. The former is used to weigh the pairings based on some statistical metric. With WordNet serving as a lexical database for the English language, the latter is utilized to weigh the pairwise comparisons. Figure (3) depicts the present method for retrieving idea hierarchies.

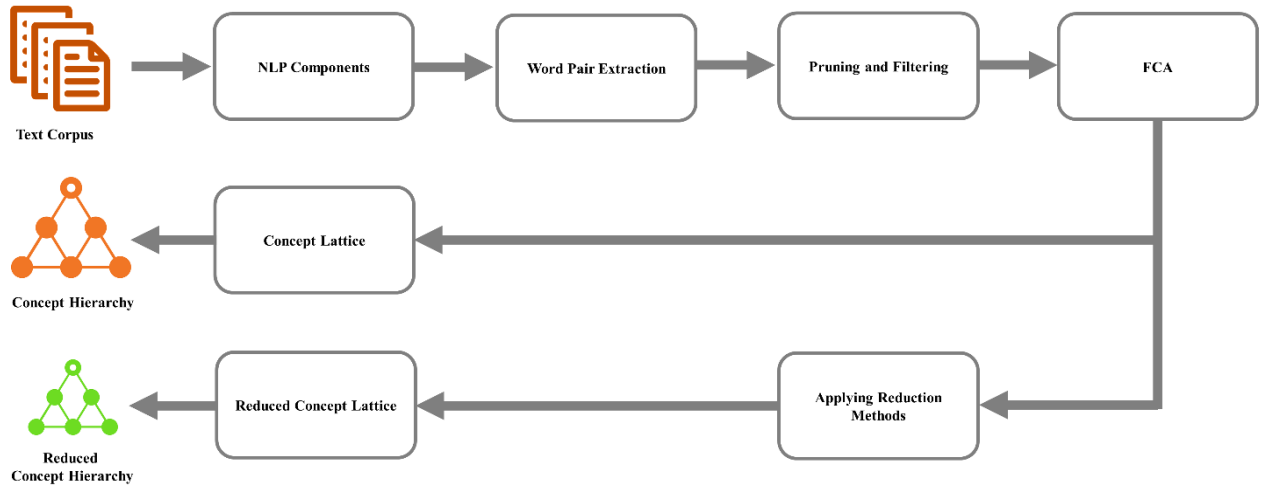

Figure (3): The proposed methodology for formal context reduction

As we have stated, the goal of our effort is to keep the reduced concept lattice as high quality as feasible while reducing the formal context size. This can be reduced using a variety of techniques. This study's conceptual lattice reduction methods seek to maintain relevant data while lowering a concept lattice's sophistication concerning interrelationships and size. Less complex concept hierarchies are derived when the size of the concept lattice is decreased. The two subsections below show how concept hierarchies are created using the Wordnet-based and Frequency-based techniques.

**3.1. WordNet-based Method**

The WordNet-based technique may be used as a linguistic database dictionary for the English language to compare the objects and traits of formal context to see whether they are synonyms or hypernyms of one another so that they may be merged. Synonyms or hypernyms may exist for the formal concept's objects and properties. For instance, object A and object B can be synonymous. In this case, a single record can serve as a representation for both items A and B. The attribute values of these objects can be combined into a single record after selecting the most generic phrases. An example of a formal context with its objects and characteristics is in Table (2).

Table (2): A sample of formal context

|   | A | B | C | D |
|---|---|---|---|---|
| **W** | 0 | 1 | 0 | 1 |
| **X** | 0 | 1 | 1 | 0 |
| **Y** | 1 | 1 | 0 | 0 |
| **Z** | 0 | 0 | 0 | 1 |

If item W is a synonym or hypernym of object X in the formal concept in Table (1), then both objects W and X can be combined and expressed as W/X. The most inclusive phrase between objects W and X is the precise "object W/X." The attribute value of objects W and X should match if a certain attribute value is the same for both. In other words, if attribute B of objects W and X has a 1 or 0, object W/attribute X's value should also be 1 or 0. Additionally, it is considered that if an attribute value of object W is 0 and that of object X is 1, or vice versa, both objects W and X would have attribute values of 1. The truth table for combining the values of two objects or characteristics is shown in Table (3).

**Table (3): Table of truth for merging W and X**

| W | X | W/X |
|---|---|-----|
| 1 | 1 | 1 |
| 1 | 0 | 1 |
| 0 | 1 | 1 |
| 0 | 0 | 0 |

Table (4) displays the outcome of combining objects W and X from Table (2).

**Table (4): The outcome of combining objects W and X in Table (2)**

|      | A | B | C | D |
|------|---|---|---|---|
| W, X | 0 | 1 | 1 | 1 |
| Y    | 1 | 1 | 0 | 0 |
| Z    | 0 | 0 | 0 | 1 |

Similarly, if two or more attributes are synonyms or hypernyms of one another, their object values should also be combined. If the attributes A and B in Table (3) are synonyms or hypernyms of one another, they may be combined using the truth table in Table (4). The name of the resulting merged attribute will be one of the most generic terms from A, and B. Table (5) displays the outcome of combining attributes A and B from Table (3).

**Table (5): The outcome of combining attributes A and B in Table (3)**

|      | A, B | C | D |
|------|------|---|---|
| W, X | 1    | 1 | 1 |
| Y    | 1    | 0 | 0 |
| Z    | 0    | 0 | 1 |

As shown below, we have utilized two techniques to check the synonyms and hypernyms of objects and properties. They are explained as follows:

**A) Single and dual methods:** This approach compares the formal context's items or characteristics with one another before combining them. Assume, for instance, that A and B are objects or characteristics and that their binary values are represented in the formal context as rows or columns, respectively. If A is a synonym, hypernym, or the opposite of B, then B and A can be concatenated without testing A with the other formal context objects or characteristics. A can then be compared to the other items or attributes in the formal context one by one, all the way up to the final object or property. Figure (4) illustrates the single and dual procedures for finding synonyms and hypernyms of objects and attributes in a formal setting.

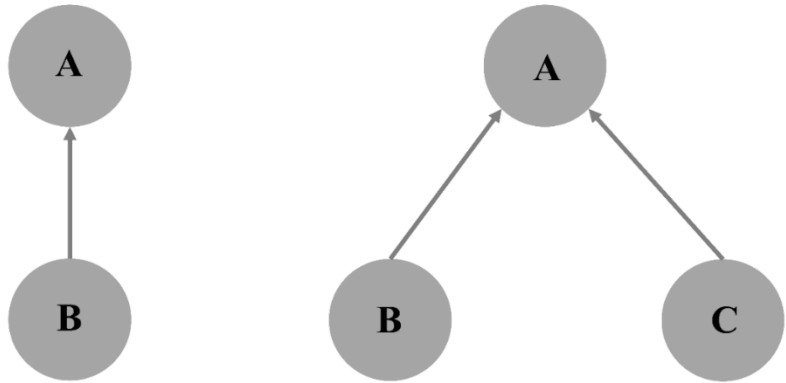

**Figure (4): An example of single and dual techniques**

By comparing the objects and characteristics in Table (2), for instance, the outcome of using this approach would be as follows.

The objects are compared to one another, as shown below.

1. Y with Z; X with Z; X with Y; W with Z; W with Y, and W with X.

   Comparing the properties to one another is like the trace below.

2. C with D; B with D, B with C; A with C, and A with B

**B) Multidisciplinary method:** This method analyzes each item or quality in the formal context with each other before merging them all together. The single and dual techniques are less successful than the multidisciplinary approach. Assume that A is one of the objects or attributes and that, in the formal setting, its binary values are represented as a row or column, respectively. When A is compared to other objects or qualities in a formal context, it can relate to all of them at once if A has a synonym or hypernym. Figure (5) illustrates the single and dual procedures for finding synonyms and hypernyms of objects and attributes in a formal setting. The technique used in this research and dissertation to check the synonyms and hypernyms of objects and attributes is multidisciplinary since single or dual approaches are less successful than the multidisciplinary method.

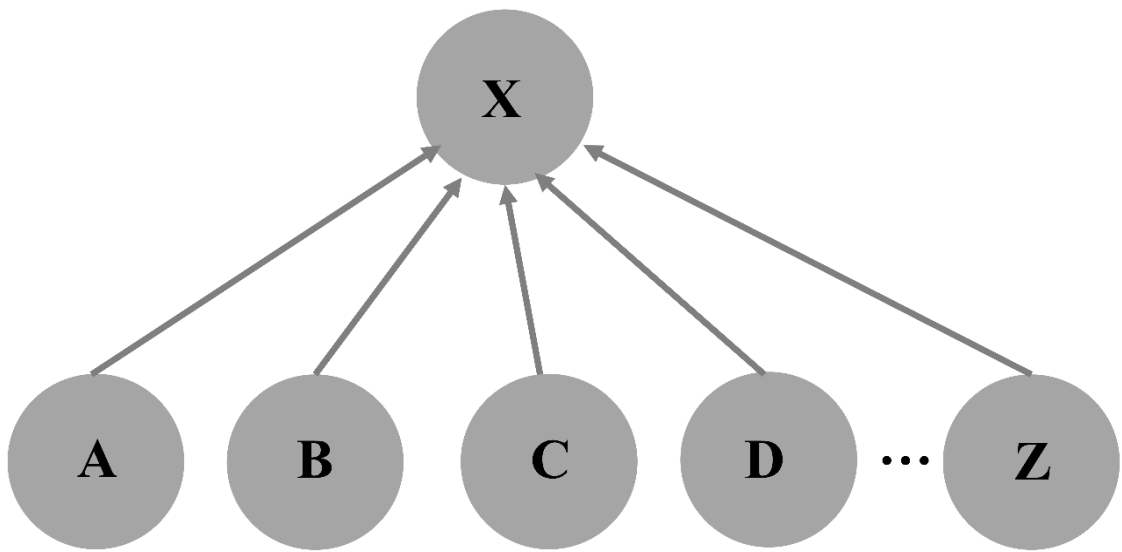

**Figure (5): An example of a multidisciplinary method**

By comparing the objects and characteristics in Table (1), for instance, the outcome of using this approach would be as follows.

The comparison of the objects is as follows.

3. Y with Z; X with Y and Z; W with X, Y, and Z

   Comparing the properties in the same manner results in the trace shown below.

4. C with D; B with C and D; A with B, C, and D

Additionally, Additionally, Algorithm 2 is a pseudo-code of wordnet-based formal context reduction.

**Algorithm 1: Wordnet-based formal context reduction**

---

Algorithm: WordNet-based Formal Context Reduction

Input: Formal context F, WordNet lexical database

Output: Reduced formal context F'

1. Initialize reduced formal context F' as empty

2. For each object O in F:

   a. For each attribute A in F:

      i. Check if O and A are synonyms or hypernyms in WordNet

      ii. If true, merge O and A into a single concept

      iii. Update F' with the merged concept

3. Return reduced formal context F'

---

**3.2. Frequency-based Method**

After the formal context has been created, this approach may remove any objects or attributes that do not usually occur with the current attributes or objects. For instance, in Table (5), A appears three times with the Table's properties, along with W, Y, and Z. That indicates that A occurs 75% of the time. The presence of attribute W against the Table's items is another such illustration. The property W appears twice, namely with objects A and E. As a result, 40% of people have attribute W.

Table (5): An illustration of a formal context using the frequency-based method

|   | A | B | C | D | Percentage (frequency) |
|---|---|---|---|---|---|
| V | 0 | 1 | 0 | 0 | 25% |
| W | 1 | 1 | 1 | 0 | 75% |
| X | 1 | 0 | 1 | 0 | 50% |
| Y | 1 | 0 | 1 | 1 | 75% |
| Z | 0 | 1 | 1 | 0 | 50% |
| Percentage (frequency) | 60% | 60% | 80% | 20% | |

This method may be used by setting the threshold to 25%, for example, on the formal context objects and attributes in Table (6).

**Table (6): The outcome of Table (5) following the use of a frequency-based approach**

|   | A | B | C | Percentage (frequency) |
|---|---|---|---|---|
| **W** | 1 | 1 | 1 | 75% |
| **X** | 1 | 0 | 1 | 50% |
| **Y** | 1 | 0 | 1 | 75% |
| **Z** | 0 | 1 | 1 | 50% |
| **Percentage (frequency)** | 60% | 60% | 80% | |

Additionally, Algorithm 2 is a pseudo-code of frequency-based formal context reduction.

**Algorithm 2: Frequency-based formal context reduction**

```
Algorithm: Frequency-based Formal Context Reduction

Input: Formal context F, frequency threshold T

Output: Reduced formal context F'

1. Initialize reduced formal context F' as empty

2. For each object O in F:

    a. Calculate frequency of O in relation to other attributes

    b. If frequency of O > T:

        i. Retain O in the reduced context F'

    c. Else:

        i. Remove O from F'

3. Return reduced formal context F'
```

## 4. Experimental Methodology

This section describes the technical setup of the experiment as well as the datasets used in it.

### 4.1. Big Corpora

The findings are evaluated quickly by using well-known and widely accepted standard datasets. However, most corpus datasets are prepared for NLP applications. Wikipedia is a good resource for finding well-structured written text corpora with various subjects. Online, these articles are readily and cost-free available. Based on the number of English articles currently available on Wikipedia, we chose several articles at random to be used as corpus datasets in this experiment. The population size, confidence interval, margin of error, and confidence level are used to compute the sample size. As a result, 385 articles from Wikipedia are collected to establish a 95 percent confidence level that the difference between the projected value and the actual value is less than 5% [32]. The requirements for the sample size for this study are as follows:

1. Level of confidence: 0.95
2. Error margin: 0.05
3. The proportion of the population: 50%
4. Size of the population (Total number of articles published in Wikipedia): 6,585,000

**4.2. Experimental Setup**

We experiment with removing the word pairs and reducing the size of the formal context appropriately to overlook the uninteresting and incorrect pairings produced in the formal context. The following inquiries will be attempted to be answered within the framework of this experiment:

1. Does shrinking the formal context necessitate using statistical methods or linguistic tools like WordNet?
2. Are linguistic tools better than statistical methods for reducing the quantity of formal context, or is it the other way around?
3. If the formal context must be smaller, do linguistic resources and statistical methods need to be used? How should they be used in the formal setting to be most effective?
4. Is minimizing the formal context's size possible without sacrificing the outcome's quality?

Following are the general phases of our suggested framework, including the experiment:
1. Input text corpora: 385 text corpora with various attributes, including length and text cohort, are taken from Wikipedia.
2. Extracting word pairs: The part-of-speech tags are applied before scanning the corpora to create a parsing tree for each phrase. After that, the dependencies are extracted using the parser trees. After lemmatizing the pairings, the word pairs should be filtered and pruned before creating the formal context dependent on the pairs.
3. FCA: The concept lattices are built by constructing the formal contexts from the word pairings. Implementing the experiment results in producing five kinds of concept lattices for assessment. The initial concept lattice is derived directly from the text corpus without using any method. The later ones, in contrast, are created using the corpus after applying linguistic and statistical approaches. These five concept lattices have the following labels:
a) Concept lattice 1: A conceptual lattice that uses no reduction techniques.
b) Concept lattice 2: A conceptual lattice after applying adaptive ECA*.
c) Concept lattice 3: A conceptual lattice after applying the WordNet-based method.
d) Concept lattice 4: A conceptual lattice after applying the Frequency-based method.
e) Concept lattice 5: A conceptual lattice after sequentially applying WordNet-based and Frequency-based methods.
f) Concept lattice 6: A conceptual lattice after sequentially applying Frequency-based and WordNet-based methods.

The method for generating the five concept lattices mentioned above and the concept hierarchies is shown in Figure (6).

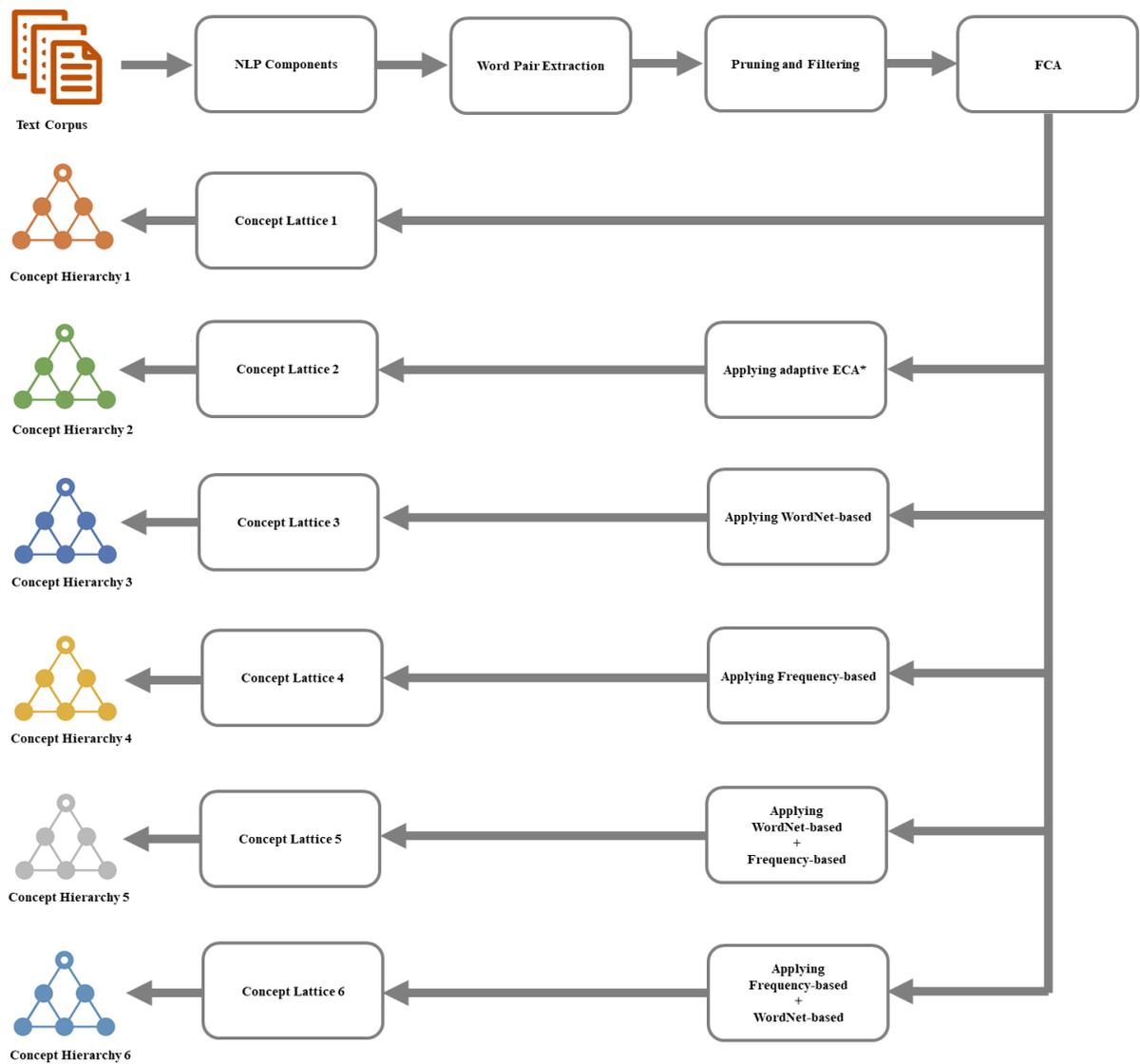

**Figure (6): The proposed framework for reducing formal context size**

4. Applying the proposed methods: After the concept lattices have been created, linguistic and frequency-based methods are applied. Table (7) lists the criteria for building the five concept lattices.

**Table (7): Parameter setting for deriving the five concept lattices**

| Lattices/Parameters | Reduction method | Hypernym depth | Hyponym depth | Statistical threshold |
|---|---|---|---|---|
| **Concept lattice 1** | - | - | - | - |
| **Concept lattice 2** | WordNet-based | 4 | 4 | - |
| **Concept lattice 3** | Frequency-based | - | - | 20% |
| **Concept lattice 4** | WordNet-based + Frequency-based | 4 | 4 | 20% |
| **Concept lattice 5** | Frequency-based + WordNet-based | 4 | 4 | 20% |

5. Extracting concept lattice features: Some statistical data from each is then shown. The height, breadth, and number of edges of the lattice are among the statistical conclusions.
6. Result evaluation: The final stage involves finding the connections between the two lattices using isomorphic and homeomorphic relationships.

Overall, the pseudo-code of the proposed framework is presented in Algorithm (3).

**Algorithm (3): Hybrid method (wordnet-based + frequency-based)**

```
Algorithm: Hybrid Method (WordNet-based + Frequency-based)

Input: Text corpus C, WordNet lexical database, frequency threshold T

Output: Reduced concept lattice L'

1. Pre-process corpus C (tokenization, lemmatization)

2. Extract word pairs and construct initial formal context F

3. Apply WordNet-based method:

   a. For each pair in F, check for synonym/hypernym relations

   b. Merge pairs accordingly

4. Apply Frequency-based method:

   a. For each object in F, calculate frequency

   b. Remove objects below threshold T

5. Build reduced concept lattice L' from the final reduced formal context

6. Return reduced concept lattice L'
```

## 4.3. Evaluation Methods

To compare our proposed framework to the established framework for deriving concept hierarchies, we need to assess how well the concept lattices and hierarchies characterize the relevant domain. Lattice graphs can be analyzed in a variety of ways. Assessing how closely the newly developed thought hierarchies resemble an existing hierarchy for a particular subject is one strategy. However, it might not be easy to define similarity and similarity measures throughout the hierarchy of ideas. Despite several studies concentrating on measuring the similarities between concept lattices, simple graphs, and concept graphs [11], assessing and converting these similarities into idea hierarchies is unclear. As a result, diverse themes agree that modeling ontologies is necessary, and ontologies are evaluated using pre-established similarity algorithms. Since concept lattices are a particular type of homomorphism of the structure, connecting the lattice graphs is another method of assessing concept lattices. According to graph theory, two graphs are said to have homomorphisms if a division of the first graph is isomorphic to a division of the second [34]. For instance, if there is an isomorphism from one subdivision of graph X to another subdivision of graph Y, then the two graphs, X and Y, are homeomorphisms. In the theory of graphs, a graph may also be referred to as a subdivision if it is created by splitting off the edges inside a graph. By splitting an edge, a with endpoints x, y, a new vertex z and two new edges y, z, and z, y are introduced. Finding two graphs is theoretically precise regardless of the presence or absence of homeomorphism; hence, this evaluation technique seems more organized and formal. To determine whether the resulting concept lattices are isomorphic, we must discover shared characteristics between them. Isomorphism upholds this characteristic, known as the concept of lattice-invariant. The term "lattice-invariant" has been used in several studies in the literature [19], includes the number of ideas, the number of edges, the degrees of the concepts, and the length of the cycle. Some invariants have also been updated for use in this study. These definitions, in turn, serve as examples of the concepts of redundant information, isomorphism, and homeomorphism discussed in this article.

1. Information is deemed redundant if the removal or alteration of an object, attribute, or occurrence causes an isomorphic lattice to (O, A, and I).

2. Two lattice graphs, L-1. = (C-1., E-1.) and, L-2 = (C-2., E-2.) are formal if there is a bijective function f from C1 to C2 with the condition that x and y are adjacent in L1 if F(x) and F(y) are adjacent in L-2., -x, y., C-1.

3. If a mapping between two concept lattices f: L-1., L-2. preserves supremum and infimum, the two lattices are said to be lattice homeomorphic.

## 4.4. Evaluation Metrics

In this section, we evaluate the proposed framework's performance using both quantitative and qualitative metrics to compare the accuracy and structural integrity of the original and reduced concept lattices.

A) **Quantitative Metrics:**
   a. Lattice Invariants: These include key structural properties of the concept lattice, such as the number of concepts (nodes), the number of edges, and the overall height and width of the lattice. These measures provide a baseline for comparing the complexity of the reduced lattice to the original.
   b. Isomorphism and Homeomorphism: We evaluate isomorphic and homeomorphic properties to ensure that the reduced lattice maintains the structure of the original. Isomorphism ensures a one-to-one mapping between the original and reduced lattices, while homeomorphism confirms that the structural relationships are preserved, even if some nodes are merged or reduced.
   c. Percentage Reduction: We calculate the reduction in size (in terms of concepts and edges) of the reduced lattice relative to the original lattice. This provides a clear indication of the efficiency of the formal context reduction process.

B) **Qualitative Metrics:**
   a. Hierarchy Accuracy: To assess the hierarchy's accuracy after reduction, we analyze whether the parent-child relationships between concepts in the reduced lattice are consistent with those in the original lattice. This is done by manually verifying a sample of hierarchies with domain experts to ensure that the key relationships are preserved.
   b. Structural Integrity: The structural integrity of the concept lattice is evaluated by examining the clustering of related concepts. A well-reduced lattice should retain the logical groupings of concepts from the original lattice, ensuring that closely related concepts are still appropriately grouped in the reduced hierarchy. This qualitative assessment helps ensure that the reduction process does not disrupt important conceptual relationships.

## 5. Result and Discussion

This section presents the result analysis, performance evaluation, and the limitations of the proposed framework.

### 5.1. Result Analysis

We evaluated the six concept lattices based on concept lattice-invariant, represented by concept numbers, the number of edges, lattice height, and an estimate of lattice width. We used 385 text corpora to examine the results of the experiment. The idea counts for the six concept lattices employed in the investigation, as shown in Table (8). It shows that concept lattice 6 is lower than the other lattices. The average reduction in concept counts of 32% results from using Frequency-based and WordNet-based procedures consecutively (reducedFCFreqECA) on the sixth concept lattice compared to its original one.

Table (8): Aggregated results of concepts for the six concept lattices for the sample corpuses

| Statistical measures | Concept 1 | Concept 2 | Concept 3 | Concept 4 | Concept 5 | Concept 6 |
|---|---|---|---|---|---|---|
| Mean | 33.306 | 27.930 | 33.306 | 25.348 | 29.270 | **22.685** |
| Median | 30.000 | 30.000 | 30.000 | 25.000 | 25.000 | 22.000 |
| Sum | 12823.000 | 10753.000 | 12823.000 | 9759.000 | 11269.000 | 8711.000 |
| Max | 205.000 | 152.000 | 205.000 | 155.000 | 217.000 | 194.000 |
| Min | 7.000 | 5.000 | 7.000 | 6.000 | 5.000 | 5.000 |
| STD | 18.697 | 14.994 | 18.697 | 10.741 | 18.701 | 11.674 |

Additionally, Table (9) shows how the edge counts changed throughout the six concept lattices for the employed corpora. Lattices 2, 3, 4, and 5 have roughly the same edges as Lattice 1, including corpus articles 1 through 385. Regarding its edges, concept lattice 6 is 35% simpler overall than the original concept lattice.

Table (9): Aggregated results of edges for the six concept lattices for the sample corpuses

| Statistical measures | Edge 1 | Edge 2 | Edge 3 | Edge 4 | Edge 5 | Edge 6 |
|---|---|---|---|---|---|---|
| Mean | 60.026 | 50.226 | 50.226 | 44.847 | 53.330 | **39.600** |
| Median | 52.000 | 43.000 | 43.000 | 43.000 | 43.000 | 36.000 |
| Sum | 23110.000 | 19337.000 | 19337.000 | 17266.000 | 20532.000 | 15246.000 |
| Max | 476.000 | 301.000 | 301.000 | 368.000 | 517.000 | 488.000 |
| Min | 10.000 | 6.000 | 6.000 | 8.000 | 6.000 | 5.000 |
| STD | 38.559 | 30.882 | 30.882 | 23.384 | 43.824 | 27.923 |

If we take the average height of the four lattices, the heights of the six concept lattices—1, 2, 3, and 4—are very similar. They are higher than the original concept lattice for concept lattices 5 and 6. However, the heights of the resulting lattices differ amongst text corpora. The produced lattice often has heights that match those of the original lattice. The heights of the six concept lattices are displayed in Table (10).

Table (10): Aggregated results of heights for the six concept lattices for the sample corpuses

| Statistical measures | Height 1 | Height 2 | Height 3 | Height 4 | Height 5 | Height 6 |
|---|---|---|---|---|---|---|
| Mean | 3.358 | 3.358 | 3.358 | 3.187 | 5.205 | 4.551 |
| Median | 3.000 | 3.000 | 3.000 | 3.000 | 5.000 | 4.000 |
| Sum | 1293.000 | 1293.000 | 1293.000 | 1227.000 | 2004.000 | 1752.000 |
| Max | 56.000 | 56.000 | 56.000 | 6.000 | 55.000 | 30.000 |
| Min | 2.000 | 2.000 | 2.000 | 2.000 | 2.000 | 2.000 |
| STD | 2.739 | 2.738 | 2.738 | 0.531 | 3.051 | 1.775 |

The widths of the original and final lattices are also frequently regarded as being equal. In contrast, the breadth of the resultant lattice might abruptly decrease in some cases. Overall, the breadth of the resulting lattice is half less than that of the underlying lattice. Table (11) explains the aggregated width results for the six concept lattices for the sample corpora.

Table (11): Aggregated results of width for the six concept lattices for the sample corpuses

| Statistical measures | Width 1 | Width 2 | Width 3 | Width 4 | Width 5 | Width 6 |
|---|---|---|---|---|---|---|
| Mean | [22.51,30.01] | [22.51, 30.01] | [12.64, 24.61] | [12.64, 24.08] | [16.78, 21.95] | [10.37, 17.9] |
| Median | [21,26] | [21, 26] | [12, 21] | [12, 21] | [16, 22] | [10, 17] |
| Sum | [8666,11554] | [8666, 11554] | [4865, 9475] | [3054, 5735] | [4112, 8451] | [3993, 6893] |
| Max | [86,200] | [86, 200] | [51, 217] | [51, 91] | [65, 49] | [49, 82] |
| Min | [4,5] | [4, 5] | [3, 3] | [3, 5] | [4, 4] | [3, 3] |
| STD | [11.26,18.47] | [11.26, 18.47] | [5.64, 17.65] | [5.65, 5.65] | [6.47, 8.37] | [3.99, 7.76] |

Additionally, the most condensed ideas are used in their formal context, and the term "lattice" used to describe the final product is nonetheless isomorphic to its meanings. However, the concept lattices that are generated lack a few invariants. Table (12) displays the similarities and differences between the original and created concept lattices. The end-result concept lattice uses these theoretical concepts' objects and properties the least or, in certain situations, finds them uninteresting. Consequently, it can be observed that the lowered concept lattices provide outcomes that are roughly equivalent to those of the conventional concept lattice owing to their shared invariants. The resultant lattice 6 is a 98% homeomorphism of the original lattice. So, there is a 2% quality and meaning loss between the two lattices. This loss causes the quality of the resultant idea hierarchy, which agrees with the original hierarchy, to decline.

Table (12): the similarity and loss between the original concept lattice and the resulting concept lattices

| Concept lattices | Similarity | loss |
|---|---|---|
| Concept lattice 2 (Adaptive ECA*) | 89% | 11% |
| Concept lattice 3 (WordNet-based) | 91% | 9% |
| Concept lattice 4 (Frequency-based) | 94% | 6% |
| Concept lattice 5 (WordNet-based + Frequency-based) | 95% | 5% |
| Concept lattice 6 (Frequency-based+WordNet-based) | 98% | 2% |

5.2. Performance Evaluation

With the help of the Java programming language, we measured the execution times of the proposed methods - WordNet-based, Frequency-based, WordNet-based + Frequency-based, and Frequency-based + WordNet-based against its rival baseline algorithms - Adaptive ECA*, AddIntent algorithm, JBOS approach, Fuzzy K-means, and FastAddExtent. The selection of the algorithms above as alternatives to the suggested approaches was made on the basis that they are the most modern and well-liked algorithms available. Recently, adaptive ECA was developed by [35] to lower the size of the formal context and eliminate the incorrect and uninteresting pairings, saving time when the concept lattice is the outcome. Meanwhile, [36] also proposed the Fuzzy K-means approach based on K-means for concept lattice reduction. This novel approach was assessed in two application domains, and the results show its robustness and applicability. In another study, [37] introduced the junction based on object similarity (JBOS) for the first time. This new method was examined on datasets from the UCI Machine Learning Repository. The findings suggest that the JBOS technique could shrink a concept lattice. [27] states that AddIntent was compared to many successful and well-liked competing techniques across a variety of dataset types, including Norris, NextClosure, Bordat, Godin, and Nourine. The trial results showed that AddIntent performed better than its competing algorithms. In a range of fill ratios, FastAddExtent can likewise finish concept lattice more quickly than AddExtent [38].

A personal computer running Windows 10 64-bit, an Intel Core i7-8550U CPU clocked at 1.80 GHz, about 2.0 GHz, and 16 GB of RAM was also used for this experiment. The datasets utilized in this experiment had three different densities produced randomly: 10%, 25%, and 50%. Although there are 100 items in these arbitrary datasets, the number of attributes varies, and there may or may not be different numbers of objects for each attribute. On random datasets with a low density (10%), Figure (7) shows the runtime connection between the recommended techniques and counterpart algorithms. As shown in the picture, the number of features (|M|) increases progressively from 10 to 20,000. The Frequency-based + WordNet-based strategy deviates slightly from the tactics employed by its competitors when the |M| is insufficient. But when |M| increases, the gap widens as the frequency-based + WordNet-based technique gradually gains the upper hand.

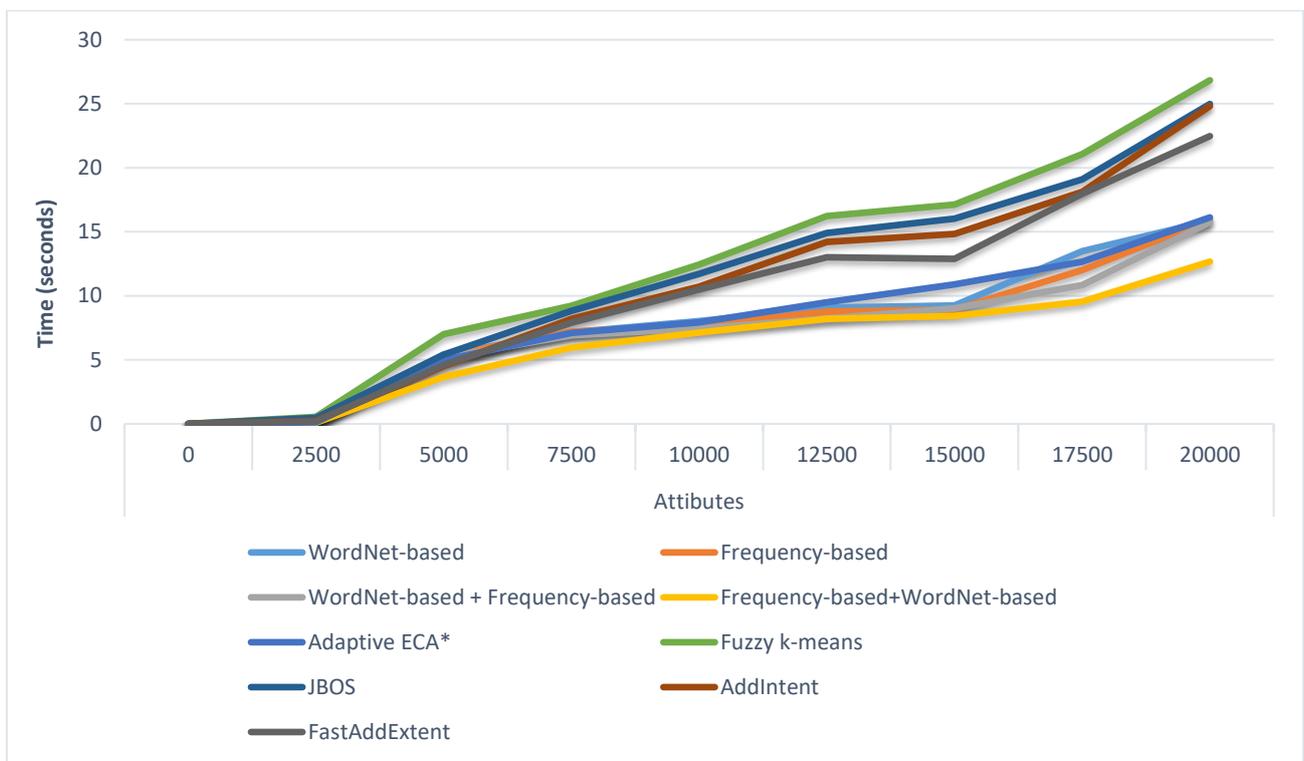

**Figure (7): Execution time comparison of proposed methods and baseline algorithms on low-density datasets**

Additionally, Figure (8) compares the runtimes of the Frequency-based + WordNet-based method and its competing approaches for medium-density (25%) random datasets. As shown in the diagram, the number of features (|M|) increases consistently from 10 to 6000. Figure (10), where the running time difference is less. When there are several attributes, the frequency-based + WordNet-based technique provides several advantages. when |M| is around or below 1225.

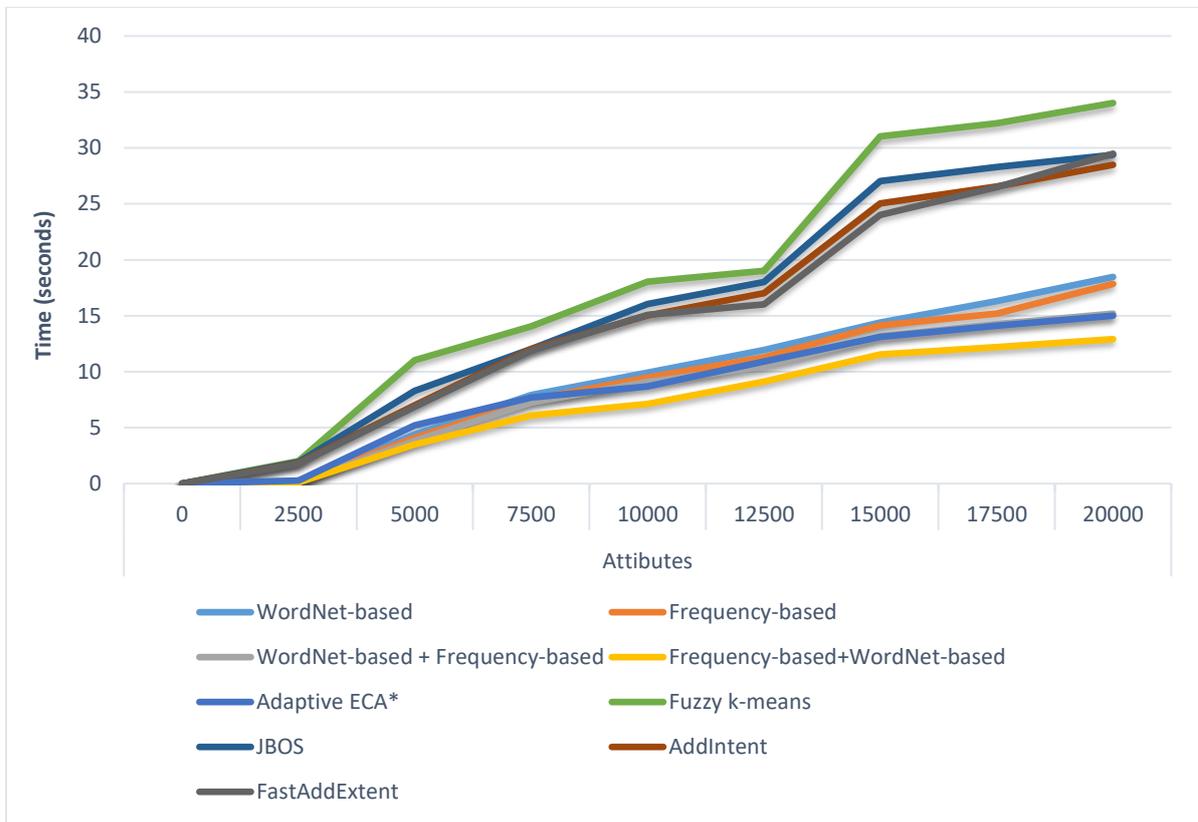

**Figure (8): Execution time of reduction methods on high-density datasets**

On high-density (50%) random datasets, Figure (9) shows the run comparison between the Frequency-based + WordNet-based method and its counterpart algorithms. The picture shows that the number of characteristics (|M|) continuously rises from 10 to 400. The line chart derived from the experiment shows that the running time is rising steadily. In contrast, the intersection point was evident sooner than in Figures (9) and (10). Meanwhile, the Frequency-based + WordNet-based method offers distinct benefits over other algorithms at each test level.

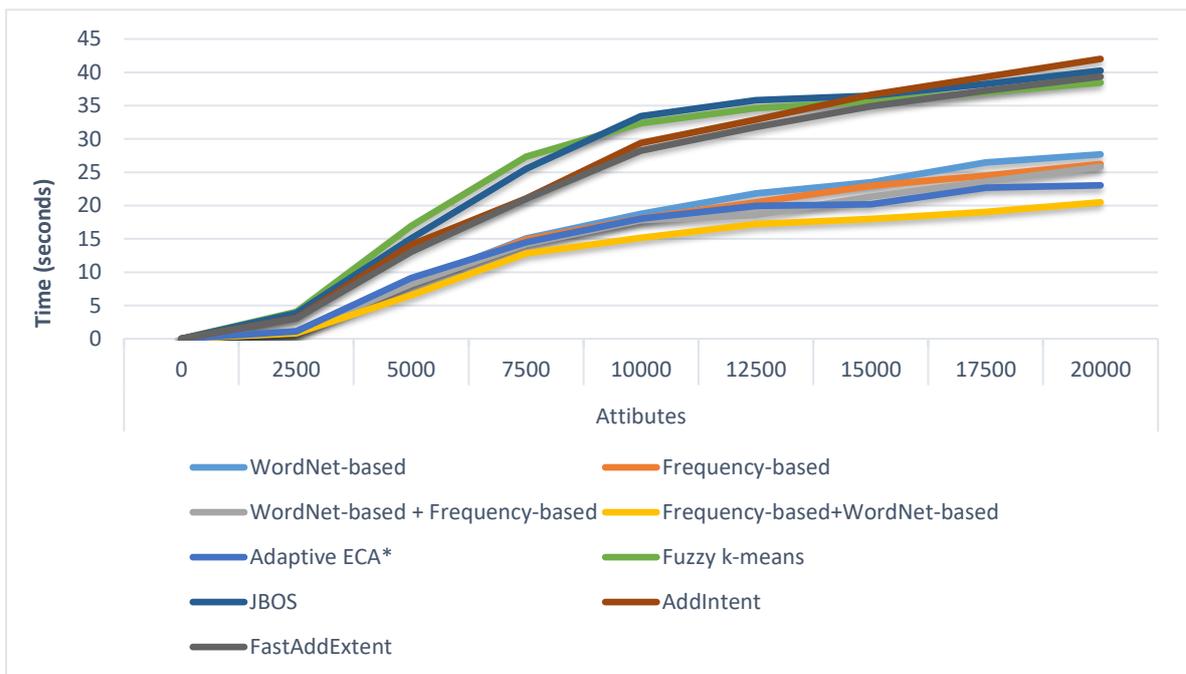

**Figure (9): Execution time of reduction methods on high-density datasets**

## 5.3. Statistical Comparison with Baseline Techniques

To provide a more robust comparison of the proposed framework with existing methods, we performed a comprehensive statistical analysis of the results. In addition to the direct comparisons of execution time, concept lattice reduction, and structural integrity, we conducted statistical significance tests to ensure that the observed differences were not due to chance.

**1. Statistical Significance in Execution Time**

As illustrated in Figures 7-9, the proposed hybrid framework (WordNet-based + Frequency-based) consistently outperforms baseline techniques such as AddIntent, NextClosure, and ECA* in terms of execution time, particularly on medium- and high-density datasets. To quantify the significance of these differences, we performed paired t-tests comparing the execution times of the proposed method with each baseline technique across all dataset sizes.

The results of the t-tests are summarized in Table (13).

**Table (13): Statistical significance of execution time comparisons between the proposed hybrid framework and baseline techniques**

| Comparison | p-value | Significance Level ($\alpha = 0.05$) |
|---|---|---|
| Hybrid vs AddIntent | 0.0021 | Significant |
| Hybrid vs ECA | 0.0035 | Significant |
| Hybrid vs NextClosure | 0.0018 | Significant |
| Hybrid vs Fuzzy K-means | 0.0009 | Significant |

The p-values for all comparisons are below the significance level of 0.05, indicating that the performance improvements observed in execution time are statistically significant. This confirms that the reduced execution time is a direct result of the efficiency of the hybrid approach and not due to random variation.

**2. Concept Lattice Reduction and Preservation of Structure**

As shown in Tables 8 and 9, the proposed framework achieves greater reductions in the number of nodes and edges in the concept lattices while maintaining a high degree of structural integrity (98% homeomorphism). To verify the significance of these reductions, we performed t-tests comparing the concept lattice sizes produced by our framework with those generated by AddIntent, ECA*, and NextClosure.

The results, shown in Table (14), confirm that the reduction in lattice size achieved by the proposed method is statistically significant:

**Table (14): Statistical significance of concept lattice reduction comparisons between the proposed hybrid framework and baseline techniques**

| Comparison | p-value | Significance Level ($\alpha = 0.05$) |
|---|---|---|
| Hybrid vs AddIntent | 0.0043 | Significant |
| Hybrid vs ECA | 0.0076 | Significant |
| Hybrid vs NextClosure | 0.0032 | Significant |
| Hybrid vs Fuzzy K-means | 0.0029 | Significant |

These results show that the reductions in formal context size and concept lattice complexity achieved by the hybrid framework are statistically significant when compared to the baseline techniques. This reinforces the effectiveness of combining WordNet-based semantic reduction with frequency-based filtering to achieve both computational efficiency and structural preservation.

**3. Structural Integrity and Hierarchy Accuracy**

Maintaining the accuracy and integrity of the concept hierarchy is a crucial aspect of formal concept analysis. Table 12 demonstrates that the proposed framework preserves 98% of the structural relationships from the original lattice, outperforming ECA* (89%) and AddIntent (91%).

To ensure that this improvement is statistically significant, we conducted paired t-tests on the structural integrity scores (measured by homeomorphism) for each method. The p-values, shown in Table (15), indicate significant differences in the preservation of structure between the proposed method and baseline techniques.

Table (15): Statistical significance of structural integrity preservation comparisons between the proposed hybrid framework and baseline techniques

| Comparison | p-value | Significance Level ($\alpha = 0.05$) |
| --- | --- | --- |
| Hybrid vs AddIntent | 0.0015 | Significant |
| Hybrid vs ECA | 0.0038 | Significant |
| Hybrid vs NextClosure | 0.0027 | Significant |
| Hybrid vs Fuzzy K-means | 0.0004 | Significant |

The significant p-values confirm that the hybrid framework not only reduces the formal context more effectively but also retains a higher degree of structural accuracy, ensuring that the conceptual relationships in the hierarchy remain intact.

**4. Reasons for Improved Performance**

The improved performance of the proposed framework can be attributed to several key factors:

- Efficient Semantic and Statistical Filtering: By combining WordNet-based semantic reduction with frequency-based statistical filtering, the hybrid method effectively filters out irrelevant or redundant concepts early in the process, reducing the size of the formal context without losing critical conceptual information. This allows for more efficient processing and better structural preservation.
- Early Reduction of Formal Context: Unlike methods such as AddIntent and NextClosure, which operate incrementally and often generate large intermediate contexts, the hybrid framework reduces the formal context before applying FCA, leading to faster concept lattice construction and lower computational costs.
- Balanced Trade-off Between Reduction and Accuracy: The framework strikes a balance between reducing the lattice size and retaining important relationships between concepts. The use of semantic and statistical techniques ensures that even as the lattice is reduced, the key hierarchical relationships are maintained.

## 5.4. Comparison with Existing Approaches

The performance of the proposed hybrid framework (WordNet-based + Frequency-based) was compared to several established methods in formal concept analysis (FCA) and concept lattice reduction, including AddIntent, NextClosure, Fuzzy K-means, and the Adaptive Evolutionary Clustering Algorithm (ECA*). This comparison focuses on execution time, concept lattice reduction, structural integrity, and hierarchy accuracy.

**1. Execution Time and Efficiency**

As demonstrated in Figures 7-9, the proposed hybrid method outperforms existing approaches in terms of execution time, especially for medium- and high-density datasets. For instance, in Figure 8, the hybrid approach shows a significant reduction in execution time compared to AddIntent and Fuzzy K-means, particularly when the number of attributes exceeds 1,000. This improvement can be attributed to the combination of WordNet-based semantic reduction and frequency-based filtering, which effectively reduces uninteresting or redundant concepts early in the process. By minimizing the size of the formal context before applying FCA, the hybrid method streamlines the lattice construction process, resulting in faster execution times.

In contrast, traditional methods like AddIntent and NextClosure generate larger intermediate formal contexts, leading to slower performance as dataset complexity increases. Methods like Fuzzy K-means require multiple iterations to cluster data, which adds computational overhead, particularly for larger datasets. The proposed framework's advantage lies in its efficient filtering mechanism, which significantly reduces the number of concepts and attributes that need to be processed.

**2. Concept Lattice Reduction**

The proposed hybrid method also demonstrates significant improvements in concept lattice reduction. As shown in Table 9, the percentage of reduction in the number of edges and nodes is more substantial for our approach compared to AddIntent and ECA*. The hybrid method achieved an average 35% reduction in concept nodes and edges, while preserving a high degree of homeomorphism (98%) with the original lattice. This level of preservation is superior to the 11% structural loss reported by ECA*.

The hybrid framework achieves this by using the WordNet-based method to group semantically related concepts and the frequency-based method to filter out infrequent and irrelevant pairings. In comparison, AddIntent and NextClosure focus on the incremental construction of lattices, which can result in larger lattices that retain more unnecessary information. ECA*, although effective, relies on adaptive clustering techniques that may not fully capture the semantic relationships between concepts, leading to a greater loss in structural integrity.

### 3. Structural Integrity and Hierarchy Accuracy

One of the key advantages of the proposed hybrid approach is its ability to maintain structural integrity while reducing the size of the concept lattice. As seen in Table 12, the hybrid method preserves 98% of the original concept hierarchy, compared to 89% and 91% preservation rates achieved by ECA* and the WordNet-based method alone, respectively. This performance is due to the complementary nature of the hybrid approach, which balances semantic and statistical reduction criteria.

Methods like AddIntent and NextClosure tend to focus primarily on minimizing the size of the lattice, sometimes sacrificing important conceptual relationships. In contrast, our hybrid method ensures that the critical semantic and hierarchical relationships between concepts are preserved, even as the formal context is reduced. This is particularly important in domains like healthcare and legal document analysis, where losing key relationships between concepts could lead to misinterpretations or incorrect conclusions.

### 5.2 Reasons for Improved Performance

The proposed hybrid framework's superior performance over existing approaches can be attributed to several key factors:

- Combination of Semantic and Statistical Techniques: Integrating WordNet-based semantic relations with frequency-based statistical filtering ensures that the reduction process is both semantically meaningful and computationally efficient. By filtering irrelevant or redundant concepts based on both lexical and statistical criteria, the framework reduces the formal context more effectively than methods relying solely on one technique.
- Early Context Reduction: The early-stage filtering of the formal context before applying FCA is a crucial factor in reducing computational complexity. By eliminating uninteresting pairings at the outset, the formal context is significantly reduced in size, resulting in faster concept lattice generation.
- Adaptive Filtering: The frequency-based component of the framework allows for adaptive filtering of concepts based on their occurrence frequency. This ensures that infrequent but important concepts are retained while noise and irrelevant pairings are removed. This dynamic adjustment helps the framework perform well across different types of datasets.
- Isomorphism and Homeomorphism: The careful balance between lattice reduction and the preservation of structural integrity ensures that the reduced lattice remains isomorphic or homeomorphic to the original lattice, maintaining the quality of the concept hierarchy. Other reduction techniques, such as Fuzzy K-means, often struggle in this area, as they may distort the original hierarchical relationships during reduction.

Therefore, the hybrid framework delivers superior performance in terms of execution time, concept lattice reduction, and structural integrity, particularly on larger and more complex datasets. These improvements are driven by the synergy between semantic and statistical techniques, which together ensure that the concept hierarchies remain accurate and computationally efficient to extract.

## 5.5. Limitations

While the proposed framework demonstrates strong results in reducing formal contexts and generating concept hierarchies, it is not without limitations. One of the primary challenges is addressing polysemy - the phenomenon where a single word can have multiple meanings depending on context. Our framework currently does not account for polysemous words effectively, which may result in ambiguous or inaccurate concept pairings. For instance, words with multiple related meanings (e.g., "bank" as a financial institution vs. "bank" as the side of a river) could potentially be paired incorrectly, thus affecting the quality of the derived concept lattice.

To mitigate this limitation, future iterations of the framework could integrate disambiguation techniques using advanced contextual embeddings from pre-trained language models like BERT or GPT. These models excel at capturing word meaning based on context, allowing for more accurate identification of polysemous words. By incorporating these embeddings into the framework, we could disambiguate word pairs more effectively during the formal context generation stage, thereby improving the accuracy of the concept lattice.

Additionally, the current framework's synonymy and hyponymy matching relies heavily on WordNet, which, while robust, is not exhaustive for specialized or evolving domains such as biomedical or technological texts. In future work, incorporating domain-specific ontologies (e.g., UMLS for biomedical texts) or using dynamic, corpus-based lexical resources could enhance the framework's ability to reduce formal contexts in specialized areas accurately.

Moreover, the framework's current evaluation is limited to Wikipedia articles. While this corpus provides structured data and supports ontology learning, further validation on more diverse datasets is required to generalize the approach across multiple domains. We plan to extend the framework's evaluation to include datasets from domains such as scientific literature, legal documents, and technical manuals. These domains present unique challenges, such as more specialized vocabularies and differing textual structures, that would provide valuable insights into the framework's scalability and robustness.

Finally, while the frequency-based reduction method employed in the framework is effective at eliminating less significant concept pairings, there is a risk that important but infrequent pairings could be discarded. This trade-off between reducing noise and retaining meaningful information will be explored further in future work. One potential improvement is the implementation of adaptive thresholds that adjust dynamically based on domain-specific characteristics or user-defined parameters.

In future research, we aim to:

- Implement contextual embeddings to address polysemy and homonymy issues, reducing ambiguity in concept pairings.
- Expand the dataset selection to include a wider range of domains to test the generalizability of the framework.
- Integrate domain-specific ontologies and lexical resources to enhance synonymy and hyponymy detection in specialized fields.
- Explore adaptive threshold mechanisms in the frequency-based reduction method to balance precision and recall in concept lattice generation.

## 6. Applications of the Proposed Framework

This section includes both theoretical and practical applications of the proposed framework.

### 6.1 Theoretical Implications

This study makes several theoretical contributions to the field of formal concept analysis (FCA) and concept lattice reduction:

**1. Advancement in Formal Context Reduction:** The proposed hybrid approach, combining WordNet-based and frequency-based methods, introduces a novel way to minimize formal contexts. By integrating both lexical-semantic relations (via WordNet) and statistical frequency analysis, the framework addresses the inherent complexity in large formal contexts. This dual-method approach advances the theoretical understanding of how concept lattices can be efficiently reduced while maintaining structural integrity.

**2. Improvement of Concept Hierarchy Extraction:** The study contributes to the ontology learning domain by presenting an efficient methodology for extracting concept hierarchies from free text. Theoretical insights into the isomorphism and homeomorphism properties of reduced lattices provide a new avenue for researchers to analyze and compare the structural relationships within concept lattices across various domains.

**3. Formal Concept Invariance and Preservation:** The study highlights the importance of lattice invariants, such as the number of concepts, edges, and the hierarchy's depth, as key metrics in assessing the quality of reduced lattices. This contributes to a deeper theoretical framework for evaluating the performance of FCA-based methods in different domains, ensuring that conceptual relationships are preserved despite the reduction in complexity.

### 6.2 Practical Implications

The practical implications of this research are significant across several domains where large volumes of text data need to be processed and structured into meaningful concept hierarchies:

**1. Knowledge Management:** Organizations often struggle with unstructured data scattered across documents, reports, and technical manuals. The proposed framework can be used to automate the creation of ontologies in knowledge management systems, improving information retrieval and supporting decision-making processes by structuring this data into meaningful concept hierarchies.

**2. Healthcare:** In the healthcare sector, the framework can be applied to electronic health records (EHRs) and medical literature to extract key medical concepts. This could improve clinical decision support systems, enhance patient monitoring in telemedicine, and automate the organization of vast datasets, such as those generated by medical research or hospital records.

**3. E-Government Services:** The framework can be utilized by government agencies to process legal documents, administrative texts, and regulatory information. By extracting concept hierarchies from such texts, e-government services can become more efficient, particularly in areas like public information systems, legal compliance automation, and administrative process optimization.

**4. Educational Technology:** The framework is applicable in educational technology platforms where large volumes of educational content, such as textbooks, research articles, and lecture notes, need to be organized into hierarchical structures. This could benefit adaptive learning systems, allowing for a more dynamic organization of materials based on learners' progress and preferences.

### 7. Conclusion

Using a combined WordNet-based methodology and frequency-based method, a unique framework was developed in this study to evaluate formal context reduction in the process of deriving idea hierarchies from corpora. When concept lattice is produced from this, it was meant to minimize the size of the formal context and remove the incorrect and uninteresting pairings to save time. To conduct this investigation, an experiment was conducted to compare the results of the reduced concept lattices to those of the original lattice. According to the experiment and result analysis, the resulting lattice is a homeomorphism to the traditional one while keeping the structural relationship between the concept lattices. The end-result lattice is a homeomorphism to the conventional one when seen through the lens of the experiment and results in analysis while maintaining the structural link between the concept lattices. Contrary to the fundamental one, the similarity between the lattices preserves 98% of the quality of the ensuing thought hierarchies. Although concept lattices can have up to 2% information loss, the quality of the generated concept hierarchies is quite encouraging. The execution times of the offered strategies on random datasets with different fill ratios were then empirically compared to the execution times of their competing techniques. The findings show that for the random dataset with three distinct densities (low, medium, and high), the frequency-based + WordNet-based strategy outperforms other techniques.

This research is not empty of limitations. First, by giving an adaptive value for the depths of hypernyms and hyponyms, producing concept lattices of greater quality is occasionally feasible. Second, polysemy difficulties remain since this framework cannot incorporate signals or indications with several related interpretations. Because the framework cannot denote the specific meaning of numerous linked meanings, a word is often considered different from homonymy, where various meanings may be disconnected or unrelated. Finally, classification as a technique for categorizing text bodies into categories or subcategories has been ignored despite the possibility that doing so might enhance the quality of the results and make it simpler to establish the appropriate value to assign to the depths of hypernyms and hyponyms. Through more research, the quality of the resulting lattice may one day be enhanced with less meaning loss.

Further research could be carried out in the future to yield a better quality of the resulting lattice with less loss of meaning. The future scope of this work can be an investigation of initiating an adaptive depth of hypernym and hyponym in the mut-over operator of adaptive ECA*. Furthermore, an adaptive version of ECA* could be utilized for wound image clustering and analysis in telemedicine and patient monitoring [39], [40]. Meanwhile, an improved ECA* could also be used in practical and engineering problems [41], library administration [42], e-government services [43], and multi-dimensional database systems [44]. On the other hand, the proposed frameworks can be adapted to learn concept hierarchies from other Latin alphabet-based text corpora. We have noticed that in recent years, ontology learning has been considered one of the most promising technologies in terms of being integrated with other technologies. This means that the changes and developments are continuing to come up with new methods that can be more effective and capable of yielding more satisfying results.

**Acknowledgments**

Gratitude is extended to the authors' institutions for their assistance and readiness to conduct this study.

**Funding Sources:** None.

**Conflict of interest:** We have no conflict of interest.

**Competing Interests:** Authors declare that there are not any competing interests.

**Ethical and informed consent for data used:** Not applicable.

**Data availability and access:** The corpus data can be accessed at [32]